\documentclass[lettersize,journal]{IEEEtran}
\usepackage{amsmath,amsfonts}
\usepackage{algorithmic}
\usepackage{algorithm}
\usepackage{array}
\usepackage[caption=false,font=normalsize,labelfont=sf,textfont=sf]{subfig}
\usepackage{textcomp}
\usepackage{stfloats}
\usepackage{url}
\usepackage{booktabs}
\usepackage{verbatim}
\usepackage{graphicx}
\usepackage{multirow}
\usepackage{cite}
\usepackage{util}
\title{Twin Trigger Generative Networks for Backdoor Attacks against Object Detection}

\begin{document}

\author{Zhiying Li, Zhi Liu, Guanggang Geng, Shreyank N Gowda, Shuyuan Lin, Jian Weng, and Xiaobo Jin
\thanks{This work is partially supported by Research Development Fund with No.
RDF-22-01-020, the top talent award project RDF-TP-0019 and National
Natural Science Foundation of China under Grant U1804159. (\textit{Corresponding author: Xiaobo Jin})}
\thanks{Zhiying Li, Zhi Liu, Guanggang Geng, Shuyuan Lin, and Jian Weng are with College of Cyber Security, Jinan,  University, Guangzhou 511436, China (email: \{tzezd2019@stu2020., gggeng@, sylin@\}jnu.edu.cn, \{peterliuforever, cryptjweng\}@gmail.com).}
\thanks{Shreyank N Gowda is with department of Computer Science, University of Nottingham, Nottingham, NG8 1BB, United Kingdom (email: shreyank.narayanagowda@nottingham.ac.uk).}
\thanks{Xiaobo Jin is with the School of Advanced Technology, Xi’an Jiaotong-Liverpool University, Suzhou 215000, China (email: xiaobo.jin@xjtlu.edu.cn).}
}

\maketitle

\begin{abstract}
Object detectors, which are widely used in real-world applications, are vulnerable to backdoor attacks. This vulnerability arises because many users rely on datasets or pre-trained models provided by third parties due to constraints on data and resources. However, most research on backdoor attacks has focused on image classification, with limited investigation into object detection. Furthermore, the triggers for most existing backdoor attacks on object detection are manually generated, requiring prior knowledge and consistent patterns between the training and inference stages. This approach makes the attacks either easy to detect or difficult to adapt to various scenarios.
To address these limitations, we propose novel twin trigger generative networks in the frequency domain to generate invisible triggers for implanting stealthy backdoors into models during training, and visible triggers for steady activation during inference, making the attack process difficult to trace.
Specifically, for the invisible trigger generative network, we deploy a Gaussian smoothing layer and a high-frequency artifact classifier to enhance the stealthiness of backdoor implantation in object detectors. For the visible trigger generative network, we design a novel alignment loss to optimize the visible triggers so that they differ from the original patterns but still align with the malicious activation behavior of the invisible triggers. 
Extensive experimental results and analyses prove the possibility of using different triggers in the training stage and the inference stage, and demonstrate the attack effectiveness of our proposed visible trigger and invisible trigger generative networks, significantly reducing the $\text{mAP}_{0.5}$ of the object detectors by 70.0\% and 84.5\%, including YOLOv5 and YOLOv7 with different settings, respectively. 


\end{abstract}

\section{Introduction}
\label{sec:intro}

Object detection is crucial in applications such as autonomous driving \cite{wu2023transformation}, \cite{zablocki2022explainability}, \cite{ma20233d}, \cite{meyer2019lasernet}, \cite{huang2023geometric} and robotic vision \cite{chan2022holocurtains}, \cite{guo2020deep}, \cite{driess2023palm}, \cite{kupcsik2021supervised}, \cite{wang2019computer}. Its widespread use in these fields highlights the need to address inherent security vulnerabilities, among which backdoor attacks have become a significant threat. The large-scale use of object detection usually relies on a large amount of training data \cite{bu2021gaia}, \cite{oquab2023dinov2}, which is a time-consuming and expensive process for dataset construction or model training for specific tasks \cite{ionescu2013human3}, \cite{lin2014microsoft}, \cite{andriluka20142d}, \cite{everingham2010pascal}. Therefore, many companies and organizations tend to use datasets or pre-trained detection models provided by third parties to save costs, but this brings security risks: the provided datasets may contain poisoned samples, or the provided pre-trained models are actually trained on poisoned data, so these possible vulnerabilities leave backdoors for attackers \cite{li2023poisoning}, \cite{huynh2024combat}, \cite{li2022backdoor}, \cite{fan2024stealthy}, \cite{ding2023backdoor}.
Once models are infected, they behave normally on clean data but behave abnormally on poisoned data, which is a serious threat to the large-scale application of object detection \cite{yin2024physical}, \cite{zhang2024detector}, \cite{cheng2023backdoor}, \cite{ma2022macab}, \cite{han2022physical}.
For example, hidden backdoor triggers may prevent detectors from identifying drivers and passengers, leading to serious safety accidents in autonomous driving. Most current research on backdoor attacks focuses on classification, however, recent research by Chan et al. \cite{chan2022baddet} and Luo et al. \cite{luo2023untargeted} reveals that backdoor attacks may be applied to object detection and highlights the urgency of addressing these evolving security challenges.

\begin{figure}[t]
    \centering
    \includegraphics[width=0.43\textwidth]{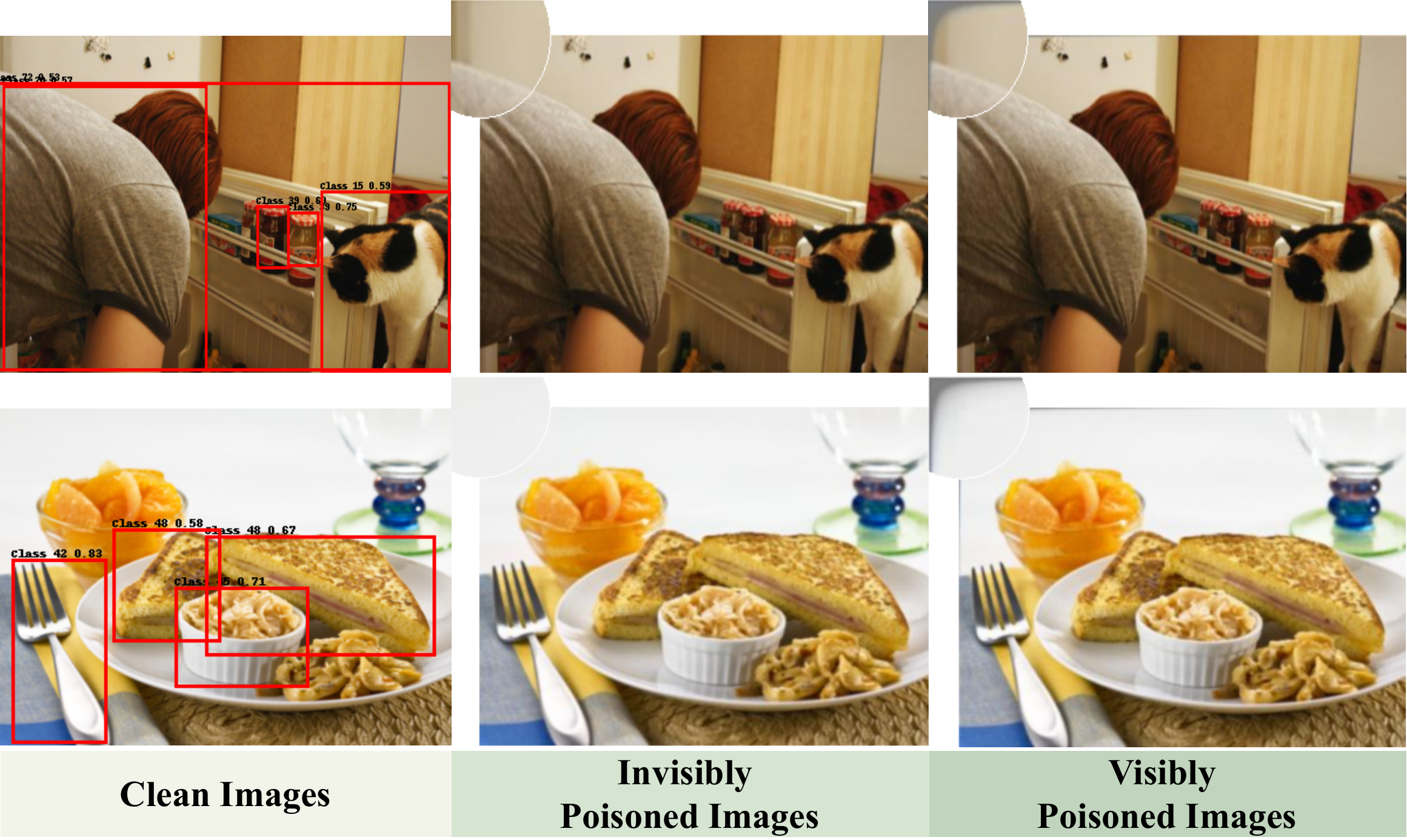}
    \caption{Output results of the victim object detector YOLOv5 constructed by our method on clean, invisible poisoned, and visible poisoned images: the detection box is output normally on the clean image, and the detection box is suppressed on the poisoned image, where the difference between the invisible and visible poisoned images is magnified in the upper left corner.}
    \label{fig:visualization}
    \vspace{-1em} 
\end{figure}

Backdoor attacks in object detection are similar to those in classification. Triggers are used to implant backdoors into object detectors during the model training stage, and these hidden backdoors are activated during the inference stage to achieve malicious purposes \cite{gu2017badnets}. Backdoor attacks can be mainly divided into two categories: visible trigger attacks and invisible trigger attacks. The former uses stable and visible patterns as triggers \cite{liu2018trojaning}, \cite{chan2022baddet}, \cite{luo2023untargeted}. These methods are relatively stable in the inference stage, can be used in the physical world, and are not easy to be removed, but are easily detected in the training stage. The latter uses techniques such as steganography and adaptive perturbations to design triggers to enhance stealthiness \cite{li2020invisible}, \cite{zhao2022defeat}, \cite{ma2022macab}. Although these methods provide better stealthiness in the training stage, they have limited adaptability and scalability in the inference stage because they can only be applied to the digital world and are easy to be removed. Current backdoor attacks on object detection face \tf{the following challenges}: 
1) Using the same trigger during training and inference greatly reduces the stealthiness of the attack, because successful detection of the trigger at any stage of the process will lead to the leakage of the only trigger pattern, causing the failure of the whole attack. 
2) The trigger is easily removed by the frequency domain classifier: Existing methods against object detection exhibit high-frequency artifacts in the frequency domain and are prone to outliers in high-frequency components, which makes the trigger very easy to detect \cite{zeng2021rethinking}.

\begin{figure*}[t]
  \centering
   \includegraphics[width=0.95\textwidth]{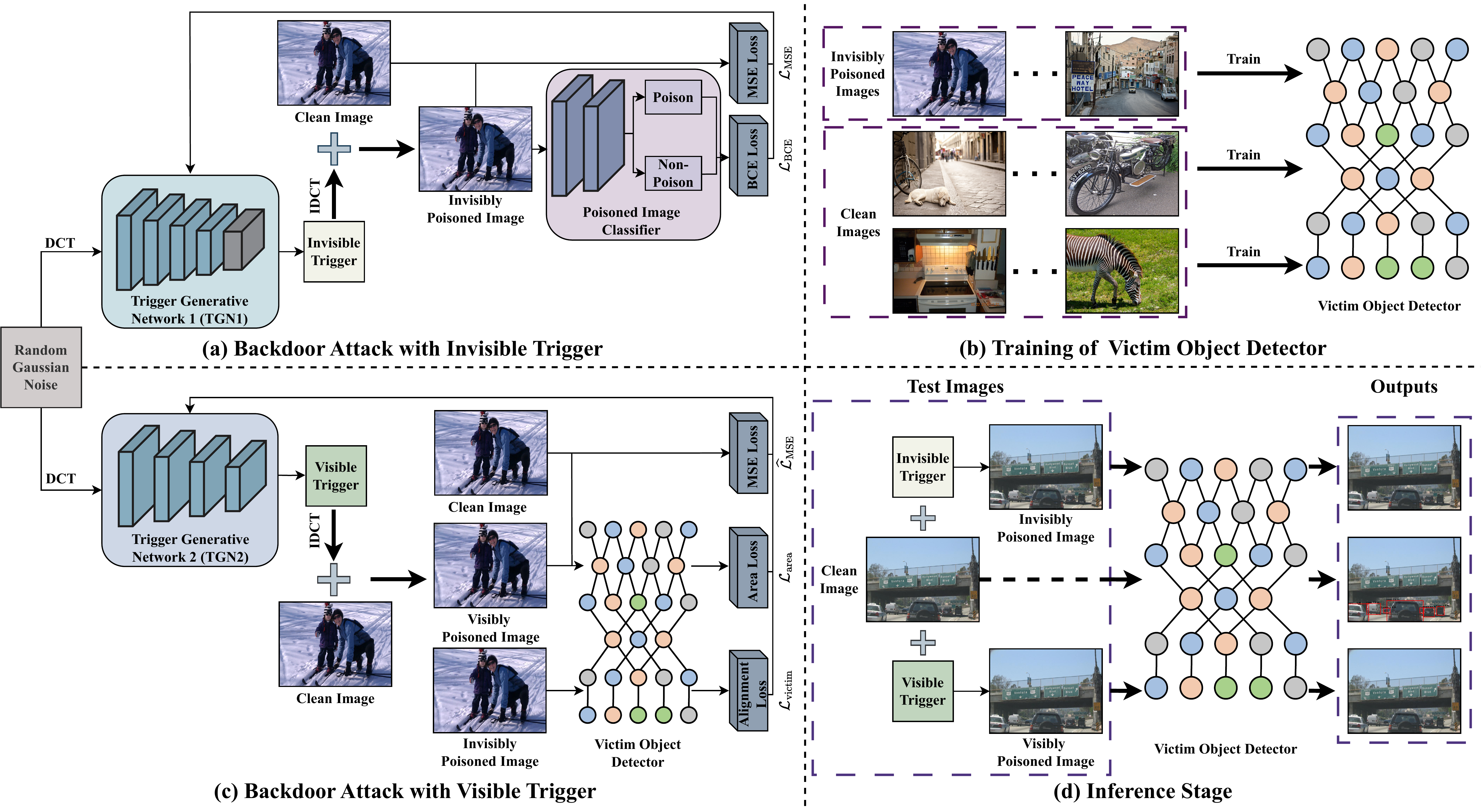}
   \caption{
   The pipeline of our work is as follows: a) A six-layer convolutional neural network and a Gaussian smoothing layer is used to generate invisible triggers in the frequency domain, where a high frequency artifacts classifier is used to enhance the stealthiness of the trigger;
   b) Both clean images and invisibly poisoned images are used to train the victim detection model; c) The visible trigger generative network generates visible triggers equivalent in behaviors to invisible triggers; d) During the inference stage, both invisibly and visibly poisoned images produce incorrect results, while clean images yield correct results.
   }
   \label{fig:pipeline}
   \vspace{-1em} 
\end{figure*}

To address these challenges, we propose twin trigger generative networks to generate invisible triggers for stealth infection and visible triggers for stable activation, as shown in Fig. \ref{fig:visualization} and Fig. \ref{fig:pipeline}. 
We first present the basic idea of constructing twin trigger generative networks through a rigorous theoretical analysis of a simple $3 \times 3$ trigger image in both frequency and spatial domains. To generate invisible triggers with better stealthiness, Trigger Generative Network 1 (TGN1) is a convolutional neural network (CNN) with a Gaussian smoothing layer and a pre-trained classifier for poisoned samples with high-frequency artifacts. To enhance the adaptability and scalability of backdoor attacks, we design Trigger Generative Network 2 (TGN2) to generate visible triggers to be consistent with the attack effect of invisible triggers generated by TGN1. Both generative networks are trained in the frequency domain to eliminate high-frequency artifacts. It is worth noting that we use invisibly poisoned images to train object detectors and use visibly poisoned images to activate hidden backdoors in the inference stage. {By comparing the Shapley value distribution in the frequency domain \cite{shapley1953value}}, the equivalence of using invisibly poisoned images and visibly poisoned images on backdoor activation can be verified. To the best of our knowledge, there is no prior research on using different triggers for the two stages. Our main contributions are as follows:

\begin{itemize}

    \item We discuss the motivation of the generative network of two types of triggers from the perspective of frequency domain and spatial domain: the visible triggers and invisible triggers in the spatial domain will be scattered and concentrated in the frequency domain. Furthermore,  we theoretically proved the equivalence of visible triggers located in the four corners of the image in the frequency domain and the feasibility of Gaussian smoothing layers for generating invisible triggers.

    \item {We propose twin trigger generative networks in the frequency domain that generate invisible triggers for implantation in training stage and visible triggers for activation in inference stage, making the attack process difficult to trace, thus achieving more stealthy backdoor attack.}

    \item Experimental results on the large-scale dataset COCO show that our method has superior attack performance compared with other visible trigger and invisible trigger attack methods, and demonstrate certain generalization in black-box attack on other detection models.
\end{itemize}

The remainder of this paper is organized as follows: Section~\ref{related} presents related work. Section~\ref{method} describes the proposed method. Section~\ref{sec:exper} presents the comprehensive experimental results of our method compared with other methods. Section~\ref{conclu} concludes the work.


\section{Related Work} \label{related}

\subsection{Visible Trigger Attacks on Image Classification.}
Gu et al. \cite{gu2017badnets} introduce BadNet, a technique that utilizes blocks of pixels as triggers. The trained victim model performs well on clean images but exhibits attacker-specific behavior when processing images with specific triggers. Chen et al. \cite{chen2017targeted} develop a method that combines natural images with hybrid attacks to enhance the effectiveness of backdoor attacks. Their results show that a high attack success rate could be achieved using approximately 50 samples. Additionally, Liu et al. \cite{liu2018trojaning} propose a Trojan attack on neural networks, achieving nearly 100\% accuracy in inducing Trojan behavior while maintaining accuracy on clean inputs. However, these methods are still easily detectable, which has led researchers to focus on making attacks more stealthy through the use of invisible triggers.

\subsection{Invisible Trigger Attacks on Image Classification.}
Li et al. \cite{li2020invisible} propose a special invisible trigger method using steganography to hide the trigger in the model, where regularization techniques are used to generate triggers with irregular shapes and sizes. Khoa Doan et al. \cite{doan2021backdoor} study triggering anomalies in latent representations of victim models and designed a triggering function to reduce representation differences. Zhao et al. \cite{zhao2022defeat} employ adaptive perturbation technology to improve the stealthiness of attacks and the flexibility of defense strategies. These methods are limited adaptability and scalability in practical scenarios where attackers can only manipulate training data. Our work explores the stealthiness and adaptability of invisible triggers in backdoor attacks on object detection from a frequency domain perspective.

\subsection{Backdoor Attacks on Object Detection.}
Backdoor attacks on object detection are an important but under-explored research area. Wu et al. \cite{wu2022just} pioneer the creation of a poisoned dataset with limited object rotation and incorrect labels. Ma et al. \cite{ma2022dangerous} use a naturalistic T-shirt as a trigger to achieve the stealthiness effect. Ma et al. \cite{ma2022macab} present model-agnostic clean-annotation backdoor (MACAB) to produce cleanly annotated images and embed a backdoor into the victim object detector in a very stealthy way. Chan et al. \cite{chan2022baddet} propose four backdoor attacks and one backdoor defense method for object detection tasks, and Luo et al. \cite{luo2023untargeted} demonstrate a simple and effective attack method in a non-targeted manner based on the task properties. Unlike these methods, we propose twin trigger generative networks to generate visible or invisible triggers instead of relying on heuristic rules.

\section{Methodology} \label{method}

\subsection{Threat Model}

\noindent \textbf{Attacker's Capacities.} In this paper, we focus on black-box attacks on object detection, which assumes that the attacker only has the ability to poison a part of the training data, but lacks any control or information about the trained model, including training losses, model architecture, and training methods. During the inference stage, the attacker can only use test images to query the trained object detector. Likewise, they do not have access to information about the model and cannot manipulate the inference process. These threats often occur when users use third-party training data, training platforms or model APIs.

\noindent \textbf{Attacker's Goals.}  Let $D = \{\{I_1,{a_1}\}, \ldots, \{I_N, {a_N}\}\}$ represents the clean dataset with $N$ images, where $I_i$ denotes the $i$-{\text{th}} image containing $n_i$ objects. Each object has an annotation as shown in the form of  $(x,y,w,h,c,p)$, where $(x,y)$ represents the left-top coordinates of the object, $w$ and $h$ denote the width and height of the bounding box respectively, $c \in \{0,1,\cdots, K - 1\}$ denotes the class label of the object and $p$ denotes the probability that the box contains an object.  Generally speaking, we can train a \tf{normal object detector} on a dataset containing only clean images. However, a backdoor attacker who overlays some clean images with a \tf{trigger} image with the same size as the clean image will obtain \tf{poisoned images}, and training on a new dataset mixed with poisonous and clean images will get \tf{victim object detector} $\widetilde{F}$, which is the \tf{target model} of backdoor attackers. Ideally, there are two main goals for the attacker: 1) For any clean image $I_i$, the victim object detector can correctly output the detection results; 2) For the poisoned image $\hat{I}_i$ synthesized from the clean image $I_i$, the victim object detector will not output any detection box.

\subsection{Visible/Invisibe Trigger in Frequency Domain} \label{ssn:frequency-domain}

The value of the image in the spatial domain is the pixel value, and its representation in the frequency domain can be obtained through discrete cosine transformation (DCT) \cite{ahmed1974discrete}, which represents the gradient magnitude of the pixel value. The DCT is a special two-dimensional discrete fourier transform (DFT), while the inverse discrete cosine transform (IDCT) is the inverse transform of DCT. Given a trigger image $G(u,v)$ with width $w$ and height $h$ in the frequency domain \cite{cooley1969fast}, its corresponding value in the spatial domain with IDCT transformation is 

\begin{equation}
\begin{split}
    I(x,y) = & \sum_{u = 0}^{w - 1}\sum_{v = 0}^{h - 1}C(u)C(v)G(u,v) \\
    & \cos\tfrac{(2x+1)u\pi }{2w}\cos\tfrac{(2y+1)v\pi }{2h}, \\
\end{split}
\end{equation}
where $x = 0,1,\ldots, w - 1$, $y = 0,1,\ldots, h - 1$, and $C(u)$ and $C(v)$ are 
\begin{small}
\begin{equation}   
    C(u)=\cs{
    \sqrt{1/w},u = 0\\
    \sqrt{2/w},u \neq 0
    }, \quad C(v)=\cs{
    \sqrt{1/h},v = 0\\
    \sqrt{2/h},v \neq 0
    }.
\end{equation}
\end{small}

Note that the coefficients $C(u)$ and $C(v)$ make the linear transformation matrix orthogonal.

In the frequency domain image, the non-zero values near the upper left corner correspond to areas with smaller gradient values in spatial domain, while the non-zero values near the lower right corner correspond to areas with larger gradient values.

In order to generate invisible triggers and visible triggers, analyses are made to explore their differences in the frequency domain and the spatial domain. In the spatial domain, we assume that the non-zero pixels in \tf{the visible trigger image} are gathered in a certain area, such as at the four corners of the image, while the non-zero pixels in \tf{the invisible trigger image} are scattered throughout the image, similar to salt and pepper noise. With the above assumption of their differences in the spatial domain, we use a binary image with size of $3 \times 3$ as a demonstration to further analyze their differences in the frequency domain. The \tf{main conclusions} are as follows:

\im{

\item When the non-zero pixels in the spatial domain are concentrated in one of the four corners, its image in the frequency domain will be spread out over the entire image.

\item When image in the frequency domain only retains the low-frequency components in the upper left corner, which can be obtained by using a Gaussian smoothing layer, the non-zero pixels in the spatial domain will scatter throughout the whole image.

\item An invisible trigger can be obtained by using a Gaussian smoothing layer because the image in the frequency domain only retains the low-frequency components in the upper left corner, causing the trigger to spread to the entire image in the spatial domain.
}

\subsubsection{Visible Trigger in Frequency Domain}

\begin{figure}[htp!]
  \centering
   \includegraphics[width=1.0\linewidth]{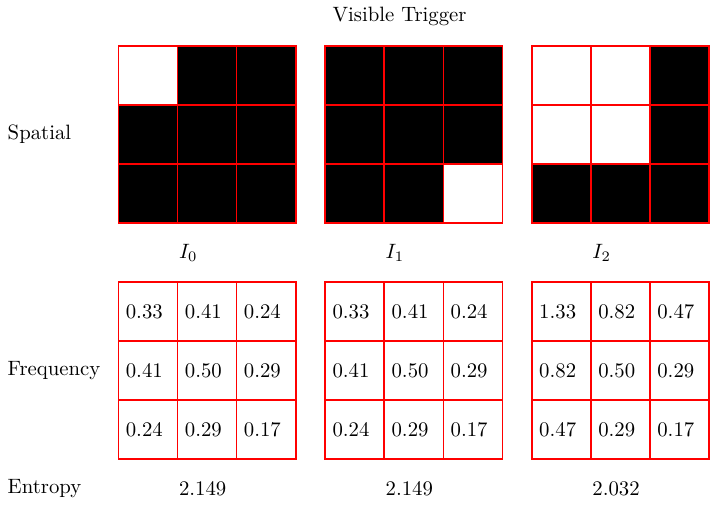}
   \caption{We place the visible trigger in the upper left corner and the lower right corner to obtain images $I_0$ and $I_1$. The visible trigger only contains one pixel, so $I_0$ and $I_1$ have the same frequency domain image. An image that is concentrated in pixels in the spatial domain will spread over the entire image in the frequency domain.}
   \label{fig:trigger}
\end{figure}

We know that for any image $I$, an image $G$ in the frequency domain will be generated after DCT transformation, and the image $G$ will be restored to an image $I$ in the spatial domain after IDCT transformation. The transformation formula is as follows (generally $w = h$)
\eqna{
G(u,v) & = & \sum_{x,y} C(u)D(v) I(x,y) f(x,y,u,v), \label{eqn:dct} \\
I(x,y) & = & \sum_{u,v} C(u)D(v) G(u,v) f(x,y,u,v), \label{eqn:idct}
}
where 
\eqn{}{
f(x,y,u,v) = \cos \tfrac{(2x + 1)u\pi}{2w} \cos \tfrac{(2y + 1)v\pi}{2h}
}
and $C(u)$ and $C(v)$ are 
\eqn{}{
C(u) = \cs{
\sqrt{1/w},u = 0\\
\sqrt{2/w},u \neq 0,
} \tm{ and } D(v) = \cs{
\sqrt{1/h},v = 0\\
\sqrt{2/h},v \neq 0.
}
}

\label{lem:cos-values}
Defining the function
\eqn{}{
h_{u}(x) = \cos \tfrac{(2x + 1)u\pi}{2w},
}
we have 
\eqn{}{
h_u(x) = h_u(w - 1 - x), \quad \tm{u is even},
}
and 
\eqn{}{
h_u(x) = -h_u(w - 1 - x),\quad \tm{u is odd}.
}

According to the assumption there is
\eqna{
h_u(w - 1 - x) & = & \cos \tfrac{(2(w - 1 - x) + 1)u\pi}{2w} \nn \\
& = & \cos \left(u\pi - \tfrac{(2x + 1)u\pi}{2w} \right) \nn \\
& = & \cos \tfrac{(2x + 1)u\pi}{2w}, \tm{u is even} \nn \\
& \tm{or} & -\cos \tfrac{(2x + 1)u\pi}{2w}, \tm{u is odd} \nn \\
& = & \pm h_u(x).
}

Therefore when $u$ is even, the function $h_u(x)$ is symmetric about the axis $x = (w - 1)/2$. Now Eqn. (\ref{eqn:dct}) and (\ref{eqn:idct}) will become
\eqna{
G(u,v) & = & \sum_{x,y} C(u)D(v) I(x,y) h_u(x)h_v(y), \label{eqn:dct} \\
I(x,y) & = & \sum_{u,v} C(u)D(v) G(u,v) h_u(x)h_v(y), \label{eqn:idct}
}
where
\eqn{}{
h_v(y) = \cos \tfrac{(2y + 1)v\pi}{2h}.
}
Similarly, the function $h_v(y)$ is symmetric about $y = (h - 1)/2$ for any even number $v$.

To facilitate discussion, we define the basic image $\mc{I}_{i,j}$ as follows
\eqn{\label{eqn:basic-img}}{
\mc{I}_{i,j}(x,y) =  \cs{
1,& x = i,y = j, \\
0,& \tm{otherwise},
}
}
and its image in the frequency domain is $\mc{G}_{i,j}$
\eqn{\label{eqn:freq-g}}{
\mc{G}_{i,j}(u,v) = C(u)D(v)h_u(i)h_v(j).
}
For any even $i$, we have
\eqn{\label{eqn:basic-freq}}{
\mc{G}_{i,j} = \mc{G}_{w - 1 - i,j}.
}
or 
\eqn{}{
\mc{G}_{i,j} = -\mc{G}_{w - 1 - i,j},
}
where $i$ is an odd number.

When one of the four corners is a trigger, as shown in Fig. \ref{fig:trigger}, for the leftmost two images $I_0$ and $I_1$ of visible triggers, we have the same frequency images $G_0$ and $G_1$
\eqn{}{
G_0 = \mc{I}_{2,0} = \mc{I}_{0,0} = \mc{I}_{0,2} = G_1.
}

Below we generalize the above conclusion to the case where the trigger includes more pixels, such as image $I_2$ contains 4 white pixels in Fig. \ref{fig:trigger}. Assume in the two trigger images $I_3$ and $I_4$, the trigger is in the top-left corner and the bottom-left corner respectively with the set of positions
\eqna{
S_3 & = & \{(0,0),(0,1),(1,0),(1,1)\} \nn \\
S_4 & = & \{(w - 1,0),(w - 1,1),(w - 2,0),(w - 2,0)\} \nn.
}

According to definition Eqn. (\ref{eqn:basic-img}), we have
\eqna{
I_3 & = & \sum_{p \in S_3} \mc{I}_{p[0],p[1]} \nn \\
I_4 & = & \sum_{p \in S_4} \mc{I}_{p[0],p[1]}. \nn 
}

At the same time, according to the additivity of DCT transformation and Eqn. (\ref{eqn:freq-g}), we obtain their frequency domain image
\eqna{
G_3(u,v) & = & \sum_{p \in S_3} \mc{G}_{p[0],p[1]}(u,v) \nn \\
& = & \sum_{p \in S_3} C(u)D(v)h_u(p[0])h_v(p[1])  \nn \\
G_4(u,v) & = & \sum_{p \in S_4} \mc{G}_{p[0],p[1]}(u,v) \nn \\
& = & \sum_{p \in S_4} C(u)D(v)h_u(w - 1 - p[0])h_v(p[1])  \nn \\
}
Therefore, fixing $v$, when $u$ is an even number, then we have
\eqn{}{
G_3(u,v) = G_4(u,v),
}
otherwise, there is 
\eqn{}{
G_3(u,v) = -G_4(u,v).
}
When $u$ is fixed, we can get similar conclusions for $v$.

\subsubsection{Invisible Trigger in Frequency Domain}
\begin{figure}[htp!]
  \centering
   \includegraphics[width=1.0\linewidth]{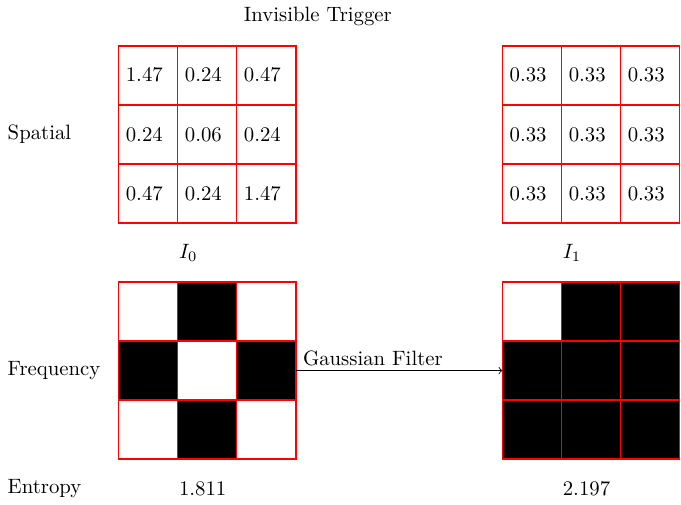}
   \caption{Image $I_0$ in the frequency domain, after passing through the Gaussian smoothing layer, its pixels from image $I_1$ in the frequency domain will be concentrated to the upper left corner and its pixels in the spatial domain will spread to the entire image.}
   \label{fig:invisible-trigger}
\end{figure}

As shown in Fig. \ref{fig:invisible-trigger}, for a noise uniformly distributed on the frequency domain image, the pixel values of the image in the spatial domain will be concentrated in the upper left corner. This phenomenon has also been verified from Fig. \ref{fig:trigger}.

In order to generate invisible triggers, we add a Gaussian smoothing layer to filter out the high-frequency component of the frequency domain image, and obtain an image (invisible trigger) evenly distributed in the spatial domain as shown in Fig. \ref{fig:invisible-trigger}. Furthermore, we found that after the Gaussian smoothing operation, the entropy of the image increased significantly.

If we define the image $\mc{G}_{i,j}$ in the frequency domain
\eqn{\label{eqn:basic-img}}{
\mc{G}_{i,j}(u,v) =  \cs{
1,& u = i,v = j, \\
0,& \tm{otherwise},
}
}
then the image $\mc{I}_{0,0}$ that contains only low frequency component and the pixel values of image $\mc{I}_{0,0}$ in the spatial domain are constant values for any $x$ and $y$
\eqn{}{
\mc{I}_{0,0}(x,y) = C(0)D(0) \cos (0)\cos (0).
}

\begin{figure*}[ht]
  \centering
   \includegraphics[width=1\linewidth]{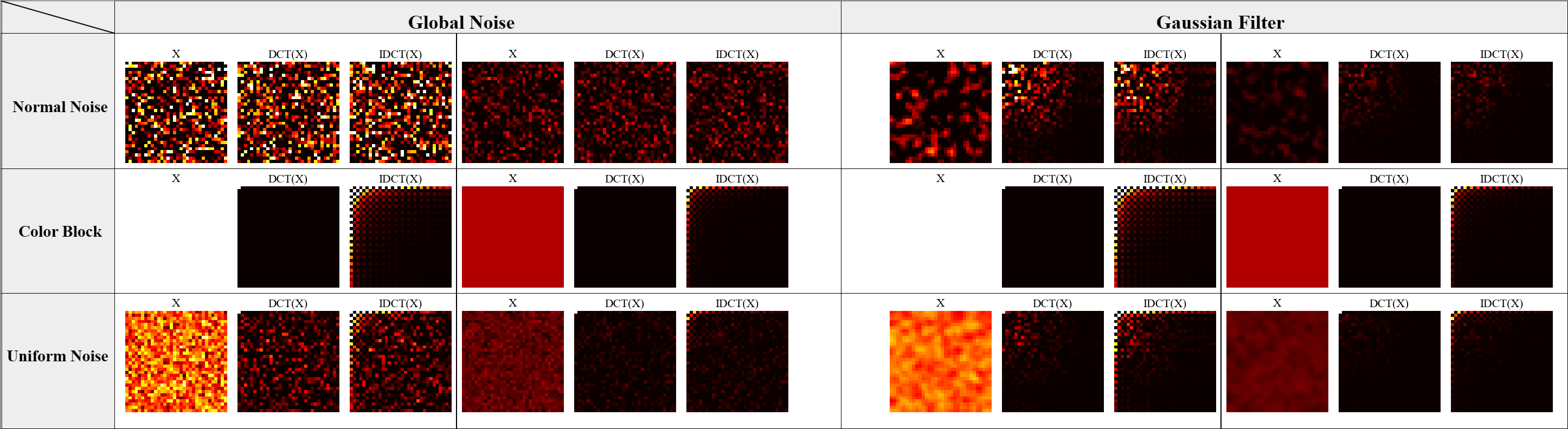}
   \caption{Correspondence between color blocks (visible), Gaussian noise and uniform noise (invisible) in the spatial domain or frequency domain: The picture on the left is before the Gaussian smoothing layer, and the picture on the right is after the Gaussian smoothing layer.}
   \label{fig:colordct}
\end{figure*}

Similarly, we can generalize to a more general situation: for an image in the frequency domain, after passing a Gaussian smoothing layer, the image values will spread to the entire image in the spatial domain. Finally, it is recommended to see Fig. \ref{fig:colordct} for more examples of color images.

Based on the above analysis, we propose the following visible trigger and invisible trigger generative network.

\subsection{Twin Trigger Generative Networks against Object Detection}


\subsubsection{Backdoor Attacks with Invisible Trigger}\label{sec:attack-invisible}

Typically, conventional invisible triggers are only effective in the spatial domain, while our goal is to create an invisible trigger that remains stealthy in both the spatial and frequency domains. To this end, we employ a Gaussian smoothing layer and a high frequency artifacts classifier, as inspired by \cite{zeng2021rethinking}, to guide the generation of the invisible trigger.

In particular, we poison the training set by random white block, random Gaussian noise, random shadow, etc., which show high-frequency artifacts phenomenon, following the configuration of \cite{zeng2021rethinking}. For each sample pair including a clean sample and its poisoned version, the clean sample is labeled as 0 and the poisoned sample is labeled as 1, thereby obtaining a class-balanced training set (equal number of samples in each class). It is worth noting that class balance is crucial for classification: when the dataset is imbalanced, the trained classifier tends to classify unknown samples into the majority class, resulting in the misclassification of samples whose ground-truth is the minority class. We then establish a high frequency artifacts classifier using a simple AlexNet \cite{krizhevsky2012imagenet} model and binary cross-entropy (BCE) loss \cite{mao2023cross}. In fact, when using Resnet50 \cite{he2016deep} instead of Alexnet, the impact on classifier performance is minimal (also see the Appendix section).

Our method utilizes a six-layer convolutional network to transform random Gaussian noise into an embedded invisible trigger. This trigger is converted from the frequency domain to the spatial domain using IDCT and is embedded into a clean image to produce poisoned samples containing an invisible trigger. As stated in previous section, images that contain only low-frequency information in the frequency domain appear as invisible white noise in the spatial domain. Therefore, we add a Gaussian smoothing layer at the end of the TGN1 to filter out high-frequency components, retaining only low-frequency ones. Consequently, the network generates invisible triggers.

Specifically, we modify the label $y_i$ of all invisibly poisoned images to $0$ to mislead the high-frequency artifact classifier and minimize the BCE loss
\begin{equation}
  \mc{L}_{\tm{BCE}} = -\frac{1}{N}\sum_{i=1}^{N}[y_{i} \log p_i + (1 - y_i) \log(1 - p_i)],
\label{eqn:bce}
\end{equation}
where $p_i$ represents the probability that the $i$-th sample is a poisoned sample. Noting that all $y_i$s are $0$, Eqn. (\ref{eqn:bce}) can essentially be simplified to
\eqn{\label{eqn:bce-probability}}{
\mc{L}_{\tm{BCE}} = - \frac{1}{N}\sum_{i = 1}^N \log (1 - p_i),
}
which is a monotonically increasing function of probability $p_i$. 

Furthermore, to ensure that invisibly poisoned images do not deviate too far from the clean images, we utilize the mean square error (MSE) loss to quantify the pixel-level difference between the clean and poisoned images 
\begin{equation}
  \mc{L}_{\tm{MSE}} = \frac{1}{N}\sum_{i = 1}^{N} \sum_{x,y} [M_i(x,y) - \widehat{M}_i(x,y)]^2,
    \label{eqn:mse}
\end{equation}
where $M_i$ and $\widehat{M}_i$ represent the clean image and the invisibly poisoned image respectively.

Finally, we optimize the parameters of the TGN1 via the following loss
\eqn{}{
\mc{L} = \mc{L}_{\tm{BCE}} + \mc{L}_{\tm{MSE}}.
}

It is worth noting that in order to make the TGN1 easier to train, we pre-trained a high frequency artifacts classifier offline on the class-balanced training set, but kept its weights fixed during the subsequent optimization process of the trigger generative network. Through the above design, it can be ensured that invisibly poisoned images are difficult to detect in the frequency domain and are visually invisible. Algorithm \ref{TGN1} outlines the training process of TGN1.

\begin{algorithm}[tb]
    \caption{Training of Trigger Generative Network1}
    \label{TGN1}
    \textbf{Input:} Random Gaussian Noise $x$ (Size: 640 $\times$ 640)\\
    \textbf{Output:} Optimized model Parameters $\theta$, Invisible Trigger $\widehat{y}$ (Size: 640 $\times$ 640) \\
    \textbf{Setting:} Epochs $T_1$ = 30, Trigger Generation Network1 (TGN1), Poisoned image classifier $p$, Clean image $M_i$, $i \in \{1, 2, \dots, N\}$, Invisibly poisoned image $\widehat{M}_i$, Learning rate $\eta$, Binary cross-entropy loss $\mathcal{L}_\text{BCE}$, Mean square error loss $\mathcal{L}_\text{MSE}$
    \begin{algorithmic}[1] 
        \STATE Initialize parameters $\theta_1$
            \FOR{$t$ = 1 to $T_1$}
            \FOR{$i$ = 1 to $N$}
                \STATE  $\widehat{y} = \text{TGN1}(x; \theta_1)$
                \STATE  $\widehat{M}_i = \widehat{y} + M_i$
                \STATE  $\mathcal{L}_1 = \mathcal{L}_\text{BCE}(p(\widehat{M}_i)) + \mathcal{L}_\text{MSE}(\widehat{M}_i, M_i)$
                \STATE \textbf{Update Parameters}
                \STATE  $\theta_1 = \theta_1 - \eta \nabla_{\theta_1} \mathcal{L}_1$  
            \ENDFOR
            \ENDFOR
            
            \STATE \textbf{return} $\theta_1$ 
    \end{algorithmic}
\end{algorithm}

\subsubsection{Training of Victim Object Detector}

We generate invisibly poisoned images (as shown in step (a) of Fig. \ref{fig:pipeline}) and clean images mixed in a certain proportion $\rho$ to form a training set. Below we describe how to train a victim object detector. Specifically, we set the width and height of the bounding boxes in the annotations of the invisibly poisoned images to zero, and the annotations of the clean images remain unchanged, thus creating a new training set. This dataset is then used to train detection models like YOLOv5 to produce a victim object detector. The detector reliably identifies objects in clean images but may fail to detect them in invisibly poisoned images. The proportion $\rho$ of poisoned samples is a crucial indicator of the attack model's effectiveness. A high attack performance with a low poisoning rate poses a significant threat to existing applications \cite{li2022backdoor}.

\subsubsection{Backdoor Attacks with Visible Trigger}

To enhance the adaptability and scalability of the attack, as shown in Fig. \ref{fig:pipeline}, TGN2 are adopted to generate the visible triggers, which is distinct from the invisible trigger generated by TGN1. During the training stage (Fig. \ref{fig:pipeline} (b)), we use invisibly poisoned images which embed invisible triggers generated by TGN1 to train the victim object detector. During the inference stage (Fig. \ref{fig:pipeline} (d)), visibly poisoned images which embed visible triggers generated by TGN2 can activate the hidden backdoor in the victim object detector equivalent to invisibly poisoned images.

Specifically, we use TGN2 to create visible triggers that are equivalent to invisible triggers with the following effects: 1) \textit{Confusion}: visible poison images are different from invisible poison images without occluding the original objects; 2) \textit{Maximization of attack effect}: visible triggers produce strong attack effects; 3) \textit{Alignment of detection results}: for images with both trigger types, the detection results on normal and victim object detectors are similar.

We use the victim object detector as a fixed API for querying, as shown in Fig. \ref{fig:pipeline} (c). TGN2 also uses six convolutional layers to generate triggers in the frequency domain, which are transformed by IDCT and combined with the clean image to synthesize visible poisoned images, where the Gaussian smoothing layer is removed to output visible triggers.

For any clean image and the visibly poisoned image, similarly, we define the following MSE loss to enhance the confusion of visibly poisoned images
\begin{equation}
  \widehat{\mc{L}}_{\tm{MSE}} = \frac{1}{N}\sum_{i = 1}^{N} \sum_{x,y} [M_i(x,y) - \widetilde{M}_i(x,y)]^2,
    \label{eqn:mse}
\end{equation}
where $M_i$ and $\widetilde{M}_i$ represent the clean image and the visibly poisoned image respectively.

Typically for a detection problem, each bounding box output by the object detector $F$ contains the following 6 parameters $(x_k,y_k,w_k,h_k,c_k,p_k)$: $(x_k, y_k)$ represents the position of the top-left corner of the bounding box, $w_k$ and $h_k$ represent the width and height of the box, and $p_k$ represents the probability that the box contains an object. Note that the target of the attack is to compress the output of the bounding box, so we do not care about the labels $c_k$ of the objects in the box. Given the object detector $F$, we propose a new function (area loss) to measure the attack effect of images on the object detector, which is the average weighted area of all bounding boxes for each image $X$ and is defined as

\eqn{\label{eqn:detection-loss-func}}{
f(X,F) = \frac{1}{n(X,F)}\sum_{k = 1}^{n(X,F)} e^{ w_k(X,F) h_k(X,F) p_k(X,F)},
}
where the object detector $F$ detects the image $X$ to output $n(X,F)$ boxes. The width, height and probability (confidence) of the $k$-th bounding box are $w_k(X,F)$, $h_k(X,F)$ and $p_k(X,F)$ respectively. Here we enhance the penalty for large area boxes through an exponential function. Therefore, we obtain the average area loss on the victim model $\widetilde{F}$ for $N$ visibly poisoned images $\widetilde{M}_i$
\eqn{}{
\mc{L}_{\tm{area}} = \frac{1}{N}\sum_{i = 1}^N f(\widetilde{M}_i,\widetilde{F}).
}

\begin{table*}[htb]
  \caption{Comparison of attack effects between our attack method and other visible trigger (Vis-Tri) and invisible trigger (Inv-Tri) attack methods on YOLOv5, where the percentage in parentheses represents the proportion of reduction of the detection metric on the clean dataset (N-Tri). The best and the second best results are {\color{red}red} and {\color{blue}blue}.}
  \label{tab:comparisons}
  \centering
  \resizebox{0.85\textwidth}{!}{
  \begin{tabular}{llrrrr}
\toprule[1.3pt]
\hline
    Attack Methods & Dataset & Precision & Recall & $\tm{mAP}_{0.5}$ & $\tm{mAP}_{0.5:0.95}$\\
    \hline
    
    Badnets \cite{gu2017badnets}  & N-Tri & 0.6210 & 0.4900 & 0.5200 & 0.3270 \\
    
      & Vis-Tri & 0.5720({\color{black}7.9\%$\downarrow $}) & 0.3980({\color{black}18.8\%$\downarrow $}) & 0.4370({\color{black}15.9\%$\downarrow $}) & 0.2720({\color{black}16.8\%$\downarrow $}) \\ 
    \cline{2-6}
    
    $l_0$\_inv \cite{li2020invisible} & N-Tri & 0.6270 & 0.4880 & 0.5220 & 0.3280 \\
     & Vis-Tri & 0.4890({\color{black}28.2\%$\downarrow $}) & 0.2880({\color{black}69.4\%$\downarrow $}) & 0.3140({\color{black}66.2\%$\downarrow $}) & 0.1910({\color{black}71.7\%$\downarrow $}) \\
    \cline{2-6}
    
    $l_2$\_inv \cite{li2020invisible} & N-Tri & 0.6570 & 0.5180 & 0.5560 & 0.3630 \\
     & Vis-Tri & {\color{blue}0.2130(67.6\%$\downarrow $)} & 0.0604({\color{black}88.3\%$\downarrow $}) & {\color{blue}0.1250(77.5\%$\downarrow $)} & {\color{blue}0.0861(76.3\%$\downarrow $)} \\
    \cline{2-6}
    
    Trojan\_sq \cite{liu2018trojaning} & N-Tri & 0.6240 & 0.4900 & 0.5200 & 0.3270 \\
     & Vis-Tri & 0.4260({\color{black}31.7\%$\downarrow $}) & 0.2240({\color{black}54.3\%$\downarrow $}) & 0.2250({\color{black}56.7\%$\downarrow $}) & 0.1360({\color{black}58.4\%$\downarrow $}) \\

    \hline

Blend \cite{chen2017targeted} & N-Tri & 0.6700 & 0.5100 & 0.5570 & 0.3620 \\
    & Inv-Tri & 0.2020({\color{black}69.8\%$\downarrow $}) & 0.0620({\color{black}87.8\%$\downarrow $}) & 0.1210({\color{black}78.3\%$\downarrow $}) & 0.0809({\color{black}77.6\%$\downarrow $}) \\
    \cline{2-6}
    
    Nature \cite{chen2017targeted} & N-Tri & 0.6400 & 0.4860 & 0.5230 & 0.3290 \\
     & Inv-Tri & 0.4480({\color{black}30.0\%$\downarrow $}) & 0.2230({\color{black}54.1\%$\downarrow $}) & 0.2270({\color{black}56.6\%$\downarrow $}) & 0.1380({\color{black}58.5\%$\downarrow $}) \\
    \cline{2-6}
    
    \hline
    
    Our Method & N-Tri & 0.6700 & 0.5220 & 0.5660 & 0.3700 \\
    on Normal Model & Inv-Tri & 0.6470({\color{black}3.4\%$\downarrow $}) & 0.5050({\color{black}3.2\%$\downarrow $}) & 0.5460({\color{black}3.5\%$\downarrow $}) & 0.3550({\color{black}4.0\%$\downarrow $}) \\
     & Vis-Tri & 0.6400({\color{black}4.5\%$\downarrow $}) & 0.5030({\color{black}3.6\%$\downarrow $}) & 0.5440({\color{black}3.9\%$\downarrow $}) & 0.3540({\color{black}4.3\%$\downarrow $})\\
    \cline{2-6}
    
    Our Method  & N-Tri & 0.6640 & 0.4980 & 0.5470 & 0.3550 \\
    on Victim Model & Inv-Tri & {\color{red}0.1680(74.7\%$\downarrow $)} & {\color{red}0.0010(99.8\%$\downarrow $)} & {\color{red}0.0846({84.5\%$\downarrow $})} & {\color{red}0.0656({81.5\%$\downarrow $})} \\
     & Vis-Tri & 0.3260({\color{black}50.9\%$\downarrow $}) & {\color{blue}0.0024(99.5\%$\downarrow $)} & 0.1640({\color{black}70.0\%$\downarrow $}) & 0.1200({\color{black}66.2\%$\downarrow $})\\
    \hline
    \bottomrule[1.0pt]
  \end{tabular}}
  \vspace{-1em} 
\end{table*}

In addition, to ensure that visibly poisoned images and invisibly poisoned images have consistent attack effects, we calculate their area losses on the victim model $\widetilde{F}$ respectively, and minimize the difference on the area losses 
\eqna{
\mc{L}_{\tm{victim}} & = & \frac{1}{N} \sum_{i = 1}^N (f(\widehat{M}_i,\widetilde{F}) - f(\widetilde{M}_i,\widetilde{F}))^2,
}
where $\widehat{M}_i$ and $\widetilde{M}_i$ represent the invisibly poisoned image and the visibly poisoned image generated by the $i$-th image $M_i$ respectively.

Finally, we optimize the parameters of the TGN2 by minimizing the following loss
\eqn{}{
\mc{L} =   \mc{L}_{\tm{area}} +  \widehat{\mc{L}}_{\tm{MSE}} 
+ \mc{L}_{\tm{victim}}.
}

\begin{algorithm}[htb!]
    \caption{Training of Trigger Generative Network2}
    \label{TGN2}
    \textbf{Input:} Random Gaussian Noise $x$ (Size: 640 $\times$ 640)\\
    \textbf{Output:} Optimized model Parameters $\theta_2$, Visible Trigger $\widetilde{y}$ (Size: 640 $\times$ 640) \\
    \textbf{Setting:} Epochs $T_2$ = 10, Trigger Generation Network2 (TGN2), Clean image $M_i$, $i \in \{1, 2, \dots, N\}$, Invisibly poisoned image $\widehat{M}_i$, Visibly poisoned image $\widetilde{M}_i$, Learning rate $\delta $, Mean square error loss $\widehat{\mc{L}}_{\tm{MSE}}$, Average area loss $\mc{L}_{\tm{area}}$, Victim loss $\mc{L}_{\tm{victim}}$
    \begin{algorithmic}[1] 
        \STATE Initialize parameters $\theta_2$
        \FOR{$t$ = 1 to $T_2$}
            \FOR{$i$ = 1 to $N$}
                \STATE  $\widetilde{y} = \text{TGN2}(x; \theta_2)$
                \STATE  $\widetilde{M}_i = \widetilde{y} + M_i$
                \STATE  
                $\mathcal{L}_2 = \widehat{\mc{L}}_{\tm{MSE}}(\widetilde{M}_i, M_i)
                + \mc{L}_{\tm{area}}(\widetilde{M}_i)$
                \STATE $\quad \quad +\mc{L}_{\tm{victim}}(\widetilde{M}_i, \widehat{M}_i)$
                \STATE  \textbf{Update Parameters}
                \STATE  $\theta_2 = \theta_2 - \delta \nabla_{\theta_2} \mathcal{L}_2$  
            \ENDFOR
            \ENDFOR
            \STATE \textbf{return} $\theta_2$ 
    \end{algorithmic}
\end{algorithm}

Algorithm \ref{TGN2} details the training process for Trigger Generative Network2 (TGN2).

\subsubsection{Inference Stage}
Finally, we synthesize poisoned images containing invisible triggers and visible triggers generated by TGN1 and TGN2 respectively to verify the attack effect of our method (as shown in Fig. \ref{fig:pipeline} (d)). Here, we will verify the following two goals of our attack system: when the clean image passes through the victim object detector, the model outputs normal bounding boxes; but when the poisoned image passes through the victim object detector, the bounding boxes output by the model will be suppressed.

\section{Experimental Analysis} \label{sec:exper}

In this section, we introduce the experimental implementation in detail and show more results on the COCO dataset \cite{lin2014microsoft}. The main hardware and software environment of the experiment is: Ubuntu 20.04.6 LTS, PyTorch library, a single NVIDIA A100 GPU and 40 GB memory.

\subsection{Implementation Details}

For object detection, we propose twin trigger generative networks in the frequency domain, which generate invisible triggers during training to implant backdoors into the object detector, and visible triggers during inference to activate them stably, rendering the attack process difficult to trace. The network architecture includes an invisible trigger generative network (TGN1) and a visible trigger generative network (TGN2), as shown in Fig. \ref{fig:pipeline}, with an input resolution of 640 × 640 pixels. The Adadelta optimizer with a dynamic learning rate (ranging from 0.001 to 0.05) is used to train the generative networks with a batch size of 64. TGN1 is trained for 30 epochs with an initial learning rate of 0.05. The learning rate is adjusted to 0.01, 0.005, and 0.001 at the 2nd, 10th, and 20th epochs, respectively. TGN2 is trained for 10 epochs with an initial learning rate of 0.01, reduced to 0.005 at epoch 5 and 0.001 at epoch 8. The victim object detector is trained according to the original training pipeline settings of the corresponding model.

\subsection{Comparison with State-of-the-art Methods}

We design a backdoor attack framework to learn visible trigger generative networks and invisible trigger generative networks respectively. Below, we compare our method with state-of-the-art visible trigger backdoor attack methods and invisible trigger backdoor attack methods respectively. These methods in Table \ref{tab:comparisons} include BadNets with a white square trigger \cite{gu2017badnets}, Trojan square (Trojan\_sq) \cite{liu2018trojaning}, hello kitty blending trigger (Blend) \cite{chen2017targeted}, nature image triggers that embed semantic information (Nature) \cite{chen2017targeted}, $l_2$ norm constraint invisible trigger ($l_2$\_inv) \cite{li2020invisible}, and $l_0$ norm constraint hidden trigger ($l_0$\_inv) \cite{li2020invisible}.  

\begin{table*}[htb!]
  \caption{Comparison of the attack effects of clean images (N-Tri), invisibly poisoned images (Inv-Tri), and visibly poisoned images (Vis-Tri) on YOLOv5 and YOLOv7 under different poisoning rates $\rho$.
  }
  \label{tab:poisoning-rate}
  \centering
  \resizebox{0.85\textwidth}{!}{
  \begin{tabular}{lrrrrrr}
  \toprule[1.3pt]
    \hline
    $\rho$ (\%) & Dataset & Model & Precision & Recall & $\tm{mAP}_{0.5}$ & $\tm{mAP}_{0.5:0.95}$\\
    \hline
    \multirow{6}*{20} & \multirow{2}*{N-Tri} & YOLOv5 & 0.6640 & 0.4980 & 0.5470 & 0.3550 \\
    ~ & ~ & YOLOv7 & 0.8040 & 0.5170 & 0.4930 & 0.3770 \\
    \cline{3-7}
    ~ & \multirow{2}*{Inv-Tri} & YOLOv5 & 0.1680 (74.70\%) & 0.0010 (99.80\%) & 0.0846 (84.53\%) & 0.0656 (81.52\%)\\
    ~ & ~ & YOLOv7 & 0.1190 (85.20\%) & 0.0005 (99.90\%) & 0.0011 (99.78\%) & 0.0009 (99.76\%) \\
    \cline{3-7}
    ~ & \multirow{2}*{Vis-Tri} & YOLOv5 & 0.3260 (50.90\%) & 0.0024 (99.52\%) & 0.1640 (70.02\%) & 0.1200 (66.20\%)\\
    ~ & ~ & YOLOv7 & 0.2370 (70.52\%) & 0.0012 (99.77\%) & 0.0024 (99.51\%) & 0.0019 (99.50\%)\\
    \hline

    \multirow{6}*{15} & \multirow{2}*{N-Tri} & YOLOv5 & 0.6530 & 0.5050 & 0.5490 & 0.3560 \\
    ~ & ~ & YOLOv7 & 0.7830 & 0.5450 & 0.5170 & 0.3960 \\
    \cline{3-7}
    
    ~ & \multirow{2}*{Inv-Tri} & YOLOv5 & 0.1490 (77.18\%) & 0.1760 (65.15\%) & 0.1310 (76.14\%) & 0.0869 (75.59\%)\\
    ~ & ~ & YOLOv7 & 0.0875 (88.83\%) & 0.0002 (99.96\%) & 0.0006 (99.88\%) & 0.0005 (99.87\%)\\
    \cline{3-7}
    ~ & \multirow{2}*{Vis-Tri} & YOLOv5 & 0.1420 (78.25\%) & 0.1720 (65.94\%) & 0.1250 (77.23\%) & 0.0823 (76.88\%) \\
    ~ & ~ & YOLOv7 & 0.2700 (65.52\%) & 0.0014 (99.74\%) & 0.0027 (99.48\%) & 0.0022 (99.44\%)  \\
    \hline

    \multirow{6}*{10} & \multirow{2}*{N-Tri} & YOLOv5 & 0.6570 & 0.5110 & 0.5550 & 0.3630 \\
    ~ & ~ & YOLOv7 & 0.7920 & 0.5460 & 0.5170 & 0.3980 \\
    \cline{3-7}
    
    ~ & \multirow{2}*{Inv-Tri} & YOLOv5 & 0.2880 (56.16\%) & 0.2380 (53.42\%) & 0.1990 (64.14\%) & 0.1290 (64.46\%)\\
    ~ & ~ & YOLOv7 & 0.3270 (58.71\%) & 0.0019 (99.65\%) & 0.0036 (99.30\%) & 0.0030 (99.25\%)\\
    \cline{3-7}
    ~ & \multirow{2}*{Vis-Tri} & YOLOv5 & 0.2390 (63.62\%) & 0.2530 (50.49\%) & 0.1950 (64.86\%) & 0.1260 (65.29\%) \\
    ~ & ~ & YOLOv7 & 0.4470 (43.56\%) & 0.0037 (99.32\%) & 0.0059 (98.86\%) & 0.0046 (98.84\%)  \\
    \hline

    \multirow{6}*{5} & \multirow{2}*{N-Tri} & YOLOv5 & 0.6620 & 0.5150 & 0.5620 & 0.3660 \\
    ~ & ~ & YOLOv7 & 0.8000 & 0.5590 & 0.5290 & 0.4060 \\
    \cline{3-7}
    
    ~ & \multirow{2}*{Inv-Tri} & YOLOv5 & 0.4890 (26.13\%) & 0.2250 (56.31\%) & 0.2470 (56.05\%) & 0.1610 (56.01\%)\\
    ~ & ~ & YOLOv7 & 0.7500 (6.25\%) & 0.0127 (97.73\%) & 0.0160 (96.98\%) & 0.0131 (96.77\%)\\
    \cline{3-7}
    ~ & \multirow{2}*{Vis-Tri} & YOLOv5 & 0.4890 (26.13\%) & 0.2240 (56.50\%) & 0.2420 (56.94\%) & 0.1570 (57.10\%) \\
    ~ & ~ & YOLOv7 & 0.7740 (3.25\%) & 0.0126 (97.75\%) & 0.0160 (96.98\%) & 0.0129 (96.82\%)  \\
    \hline

    \multirow{6}*{1} & \multirow{2}*{N-Tri} & YOLOv5 & 0.6710 & 0.5240 & 0.5660 & 0.3700 \\
    ~ & ~ & YOLOv7 & 0.7890 & 0.5640 & 0.5330 & 0.4120 \\
    \cline{3-7}
    
    ~ & \multirow{2}*{Inv-Tri} & YOLOv5 & 0.6560 (2.24\%) & 0.5030 (4.01\%) & 0.5450 (3.71\%) & 0.3540 (4.32\%)\\
    ~ & ~ & YOLOv7 & 0.8280 (-4.94\%) & 0.4970 (11.88\%) & 0.4760 (10.69\%) & 0.3680 (10.68\%)\\
    \cline{3-7}
    ~ & \multirow{2}*{Vis-Tri} & YOLOv5 & 0.6550 (2.38\%) & 0.4980 (4.96\%) & 0.5430 (4.06\%) & 0.3520 (4.86\%) \\
    ~ & ~ & YOLOv7 & 0.8240 (-4.44\%) & 0.4940 (12.41\%) & 0.4730 (11.26\%) & 0.3660 (11.17\%)  \\
    \hline

    \multirow{6}*{0.5} & \multirow{2}*{N-Tri} & YOLOv5 & 0.6440 & 0.5200 & 0.5570 & 0.3620 \\
    ~ & ~ & YOLOv7 & 0.7930 & 0.5290 & 0.5040 & 0.3860 \\
    \cline{3-7}
    
    ~ & \multirow{2}*{Inv-Tri} & YOLOv5 & 0.6470 (-0.47\%) & 0.4870 (6.35\%) & 0.5360 (3.77\%) & 0.3460 (4.42\%)\\
    ~ & ~ & YOLOv7 & 0.8150 (-2.77\%) & 0.4770 (9.83\%) & 0.4580 (9.13\%) & 0.3510 (9.07\%)\\
    \cline{3-7}
    ~ & \multirow{2}*{Vis-Tri} & YOLOv5 & 0.6610 (-2.64\%) & 0.4870 (6.35\%) & 0.5350 (3.95\%) & 0.3460 (4.42\%) \\
    ~ & ~ & YOLOv7 & 0.8110 (-2.27\%) & 0.4750 (10.21\%) & 0.4560 (9.52\%) & 0.3490 (9.59\%)  \\
    \hline

    \multirow{6}*{0} & \multirow{2}*{N-Tri} & YOLOv5 & 0.6700 & 0.5220 & 0.5660 & 0.3700 \\
    ~ & ~ & YOLOv7 & 0.8070 & 0.5680 & 0.5410 & 0.4170 \\
    \cline{3-7}
    ~ & \multirow{2}*{Inv-Tri} & YOLOv5 & 0.6470 (3.43\%) & 0.5050 (3.26\%) & 0.5460 (3.53\%) & 0.3550 (4.05\%)  \\
    ~ & ~ & YOLOv7 & 0.8310 (-2.97\%) & 0.5140 (9.51\%) & 0.4930 (8.87\%) & 0.3810 (8.63\%)\\
    \cline{3-7}
    ~ & \multirow{2}*{Vis-Tri} & YOLOv5 & 0.6400 (4.48\%) & 0.5030 (3.64\%) & 0.5440 (3.89\%) & 0.3540 (4.32\%) \\
    ~ & ~ & YOLOv7 & 0.8280 (-2.60\%) & 0.5070 (10.74\%) & 0.4860 (10.17\%) & 0.3750 (10.07\%)\\
    \hline
    \bottomrule[1.0pt]
  \end{tabular}}

\end{table*}

We construct poisoned datasets using different attack methods, with a training set poisoning rate of $\rho=20\%$, consisting of 20\% poisoned and 80\% clean images. As shown in Table \ref{tab:comparisons}, we employ YOLOv5 \cite{yao2022detection} as the object detector to test three datasets (clean dataset without trigger (N-Tri), datasets containing invisible trigger (Inv-Tri) and visible trigger (Vis-Tri)).
Notably: 1) We only use invisible triggers to train YOLOv5 in our method, but use both triggers during testing; 2) We evaluate the performance of both triggers under the normal and the victim object detectors.

Table \ref{tab:comparisons} shows detection metrics for each method on datasets: Precision, Recall, $\tm{mAP}_{0.5}$, and $\tm{mAP}_{0.5:0.95}$ \cite{chan2022baddet, wang2023yolov7}. The percentages in brackets indicate the performance degradation on the poisoned dataset with the corresponding trigger compared to the clean dataset. A greater reduction ratio indicates a stronger attack effect. Analyzing the results in Table \ref{tab:comparisons} yields three main conclusions:
\im{

\item The dataset with invisible triggers generated by our method shows the largest performance drop in all metrics on the victim object detector: the $\tm{mAP}_{0.5}$ and $\tm{mAP}_{0.5:0.95}$ decrease by $84.5\%$ and $81.5\%$, surpassing the biggest rivals' $77.5\%$ and $76.3\%$, respectively.

\item Unlike previous methods that use visible triggers, our method attacks a victim model trained with invisible triggers, resulting in a weaker attack effect. However, it still achieves a $\tm{mAP}_{0.5}$ reduction of $70.0\%$. The setup of our method is closer to real-life scenarios, as images with invisible triggers are less perceptible and easier to blend with clean samples. Poisoned images with visible triggers are more easily constructed by attackers.

\item The poisoned images (including visible triggers or invisible triggers) generated by our method have limited impact on the normal object detector, causing only slight decreases in detection performance.
}

\subsection{Impact of Poisoning Rate}

Table \ref{tab:poisoning-rate} shows the comparison of the attack effects of poisonous samples on the YOLOv5 \cite{yao2022detection} and YOLOv7 \cite{wang2023yolov7}
for the victim models trained with different poisoning rates, where percentages in parentheses indicate the proportion of degradation in detection performance.

It can be seen that when the poisoning rate reaches 20\%, the attack effect of our trigger generative networks on the attacked model YOLOv7, whether visible or invisible trigger, reaches almost 100\%. Even for the normal object detector YOLOv7 (poisoning rate $\rho = 0$), when we use our method to attack, there is a loss of 8\%-10\% in detection performance. Comparing the two victim models YOLOv5 and YOLOv7, it is found that YOLOv7 is more vulnerable to attack at all different poisoning rates.  At the same time, we also found that the detection performance of poisoned samples on the normal model has increased, which also verifies that YOLOv7 has strong feature extraction capabilities and is more sensitive to triggers (noise) in images.

\subsection{Ablation Study on Losses}

We verified the role of different loss functions in training the invisible trigger generative network and the visible trigger generative network respectively, as shown in Table \ref{tab:ablation-trigger}.

\begin{table}[t]
    \caption{Ablation study on losses in visible/invisible trigger generative network.} 
    \label{tab:ablation-trigger}
    
    \resizebox{0.45\textwidth}{!}{
    \centering
    \begin{tabular}{lrrrr}
    \toprule[1.3pt]
    \multicolumn{5}{c}{Visible Trigger} \\
    \midrule
    Losses & Precision & Recall & $\text{mAP}_{0.5}$ & $\text{mAP}_{0.5:0.95}$ \\
    \midrule
    $\mathcal{L}_{\text{area}}$ & 0.1250 & 0.2270 & 0.1380 & 0.0901 \\ 
    + $\widehat{\mathcal{L}}_{\text{MSE}}$ &  0.1210 & 0.2380 & 0.1410 & 0.0918 \\
    + $\mathcal{L}_{\text{victim}}$ & \textbf{0.1250} & \textbf{0.2180} & \textbf{0.1360} & \textbf{0.0889} \\
    \bottomrule[1.0pt]
    \end{tabular}
    }
    
    \vspace{1.5pt}
    
    \resizebox{0.45\textwidth}{!}{
    \centering
    \begin{tabular}{lrrrr}
    \toprule[1.3pt]
    \multicolumn{5}{c}{Invisible Trigger} \\
    \midrule
    Losses & Precision & Recall & $\text{mAP}_{0.5}$ & $\text{mAP}_{0.5:0.95}$ \\
    \midrule
    $\mathcal{L}_{\text{MSE}}$ & 0.2110 & 0.2150 & 0.1800 & 0.1220 \\
    $\mathcal{L}_{\text{BCE}}$ & 0.2100 & 0.1580 & 0.1580 & 0.1060 \\
    $\mathcal{L}_{\text{MSE}}$ + $\mathcal{L}_{\text{BCE}}$ & \textbf{0.1460} & \textbf{0.1190} & \textbf{0.1120} & \textbf{0.0742} \\
    \bottomrule[1.0pt]
    \end{tabular}
    }
  
\end{table}

As can be seen from Table \ref{tab:ablation-trigger}, in the invisible trigger generative network, $\mc{L}_{\tm{BCE}}$ plays a greater role than $\mc{L}_{\tm{MSE}}$, but it is obvious that using both at the same time has the best effect.
In the visible trigger generative network, the role of loss MSE $\mc{L}_{\tm{MSE}}$ is almost negligible, because the main goal of the method is to generate visible triggers that are consistent with the effect of invisible triggers, so $\mc{L}_{\tm{area}}$ and $\mc{L}_{\tm{victim}}$ play a more important role.

\begin{table}[htb!]
\caption{Comparison of the poisoned image classifier effects of poisoned samples generated by various attacked methods on Alexnet and Resnet50.}
  \label{tab:resnet50}
  \centering
  \resizebox{\linewidth}{!}{
  \begin{tabular}{lrrrrr}
  \toprule[1.3pt]
    \hline
    Attack Methods & Model & Accuracy & Precision & Recall & F1 Score\\
    \hline

    \multirow{2}*{Badnets} & Alexnet & 0.9194 & 0.9656 & 0.8703 & 0.9148 \\
    ~ & Resnet50 & 0.9404 & 0.9422 & 0.9389 & 0.9402 \\
    \cline{2-6}

    \multirow{2}*{$l_0$\_inv} & Alexnet & 0.5000 & 0.4336 & 0.0314 & 0.0580 \\
    ~ & Resnet50 & 0.4986 & 0.4658 & 0.0554 & 0.0976 \\
    \cline{2-6}

    \multirow{2}*{$l_2$\_inv} & Alexnet & 0.6587 & 0.9185 & 0.3489 & 0.5020 \\
    ~ & Resnet50 & 0.6397 & 0.8533 & 0.3376 & 0.4798 \\
    \cline{2-6}

    \multirow{2}*{Trojan\_{sq}} & Alexnet & 0.5188 & 0.6821 & 0.0690 & 0.1236 \\
    ~ & Resnet50 & 0.5069 & 0.5588 & 0.0720 & 0.1255 \\
    \cline{2-6}
    
    \multirow{2}*{Trojan\_{wm}} & Alexnet & 0.5769 & 0.8581 & 0.1853 & 0.3011 \\
    ~ & Resnet & 0.5489 & 0.7246 & 0.1559 & 0.2527 \\
    \cline{2-6}

    \multirow{2}*{Blend} & Alexnet & 0.7660 & 0.9480 & 0.5635 & 0.7041 \\
    ~ & Resnet50 & 0.8156 & 0.9230 & 0.6893 & 0.7877 \\
    \cline{2-6}
    
    \multirow{2}*{Nature} & Alexnet & 0.5666 & 0.8450 & 0.1646 & 0.2722 \\
    ~ & Resnet50 & 0.5740 & 0.7754 & 0.2061 & 0.3217 \\
    \hline
    
    \multirow{2}*{Average Measure} & Alexnet & 0.6438 & 0.8073 & 0.3190 & 0.4108 \\
    ~ & Resnet50 & 0.6463 & 0.7490 & 0.3507 & 0.4293 \\
    \cline{2-6}

    \hline
    \bottomrule[1.0pt]
  \end{tabular} }
  
\end{table}
\subsection{Ablation Study on Poison Samples Classifier}\label{ssn:classifier}
In our pipeline, we choose to use the simplest and most efficient Alexnet \cite{krizhevsky2012imagenet} as the high-frequency artifact classifier (Poisoned Image Classifier in Fig. \ref{fig:pipeline} (a)). At the same time, we also use the more common Resnet50 \cite{he2016deep} as the high-frequency artifact classifier instead of Alexnet. The comparison results are shown in Table \ref{tab:resnet50}. In general, Resnet50 has comparable performance to Alexnet on multiple classification measurement, and the average performance difference of F1 performance measure is close to 2 points.

\subsection{Trigger Transferability on Victim Models} \label{ssn:transferability}

Table \ref{tab:transferability} shows the transferability of our twin trigger generative networks across multiple detectors (YOLOv5, YOLOv7, Faster R-CNN \cite{ren2015faster}). We use them as the victim object detector in training stage, then test the attack effect of twin trigger generative networks. The results suggest that: 1) Our method has better attack effects on YOLOv7 and Faster R-CNN than YOLOv5; 2) Our method has strong scalability and can attack different detectors.

\begin{table}[t]
  \centering
  \caption{Transferability of our method across YOLOv5, YOLOv7, and Faster R-CNN (F-RCNN), with a poisoning rate of 20\%. The percentages in parentheses indicate the performance reduction on the clean dataset (N-Tri) when using invisible triggers (Inv-Tri) or visible triggers (Vis-Tri).}
\label{tab:transferability}
  \resizebox{0.49\textwidth}{!}{
  \begin{tabular}{lcrrrr}
  \toprule[1.3pt]
    \hline
    Model & Dataset & Precision & Recall & $\tm{mAP}_{0.5}$ & $\tm{mAP}_{0.5:0.95}$\\
    \hline
     \multirow{5}*{YOLOv5} & N-Tri & 0.6640 & 0.4980 & 0.5470 & 0.3550 \\
\cline{3-6}
     ~ & \multirow{2}*{Inv-Tri} & 0.1680 & 0.0010  & 0.0846 & 0.0656 \\
    ~ & ~ & (74.70\%$\downarrow $) & (99.80\%$\downarrow $) & (84.53\%$\downarrow $) & (81.52\%$\downarrow $) \\
\cline{3-6}
   ~ & \multirow{2}*{Vis-Tri} & 0.3260 & 0.0024 & 0.1640 & 0.1200 \\
    ~ & ~ & (50.90\%$\downarrow $) & (99.52\%$\downarrow $) & (70.02\%$\downarrow $) & (66.20\%$\downarrow $) \\

\hline

    \multirow{5}*{YOLOv7} & N-Tri & 0.8040 & 0.5170 & 0.4930 & 0.3770 \\
\cline{3-6}
    ~ & \multirow{2}*{Inv-Tri} & 0.1190  & 0.0005  & 0.0011  & 0.0009 \\
     ~ & ~ & (85.20\%$\downarrow $) & (99.90\%$\downarrow $) & (99.78\%$\downarrow $) & (99.76\%$\downarrow $) \\
\cline{3-6}
     ~ & \multirow{2}*{Vis-Tri} & 0.2370  & 0.0012  & 0.0024  & 0.0019 \\
     ~ & ~ & (70.52\%$\downarrow $) & (99.77\%$\downarrow $) & (99.51\%$\downarrow $) & (99.50\%$\downarrow $) \\

\hline

    \multirow{5}*{F-RCNN} & N-Tri & 0.4594 & 0.2480 & 0.4590 & 0.2820 \\
\cline{3-6}
    ~ & \multirow{2}*{Inv-Tri}  & 0.0305  &  0.0270  & 0.0310  & 0.0190  \\
     ~ & ~ & (93.36\%$\downarrow $) & (89.11\%$\downarrow $) & (93.25\%$\downarrow $) & (93.26\%$\downarrow $) \\
\cline{3-6}
    ~ & \multirow{2}*{Vis-Tri}  & 0.0311  &  0.0280  & 0.0310  & 0.0190 \\
     ~ & ~ & (93.23\%$\downarrow $) & (88.71\%$\downarrow $) & (93.25\%$\downarrow $) & (93.26\%$\downarrow $) \\
\hline
\bottomrule[1.0pt]
  \end{tabular}}
  \vspace{-1em} 
\end{table}

\subsection{Equivalence of Visible and Invisible Triggers}

\begin{figure}[t]
    \centering
    \includegraphics[width=0.35\textwidth]{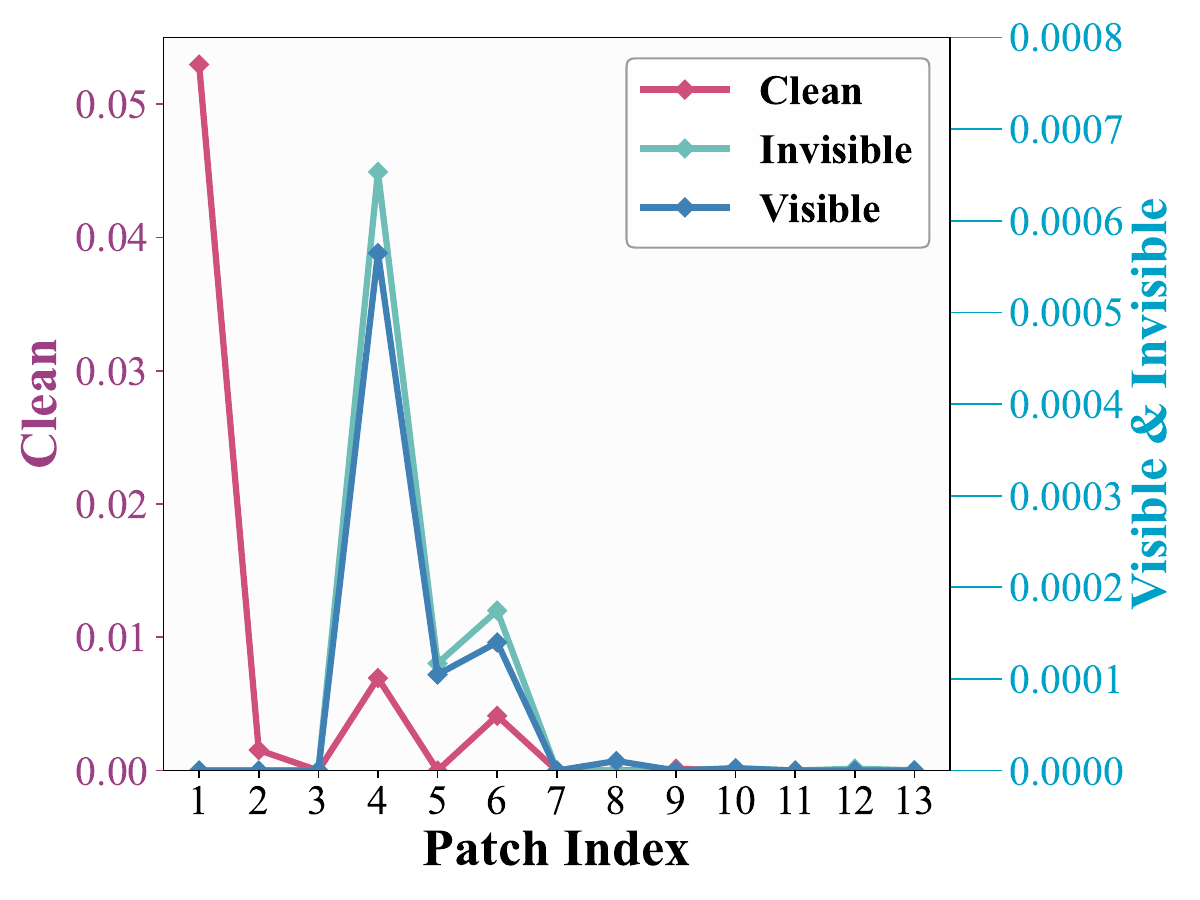}
    \caption{Comparisons of the Shapley values of clean, invisibly poisoned, and visibly poisoned images, where the first and last points of each curve represent the contribution of the lowest and highest frequency part, respectively.}
    \label{fig:shapley-value}
    \vspace{-1.2em} 
\end{figure}

\begin{figure*}[t]
  \centering
   \includegraphics[width=1\linewidth]{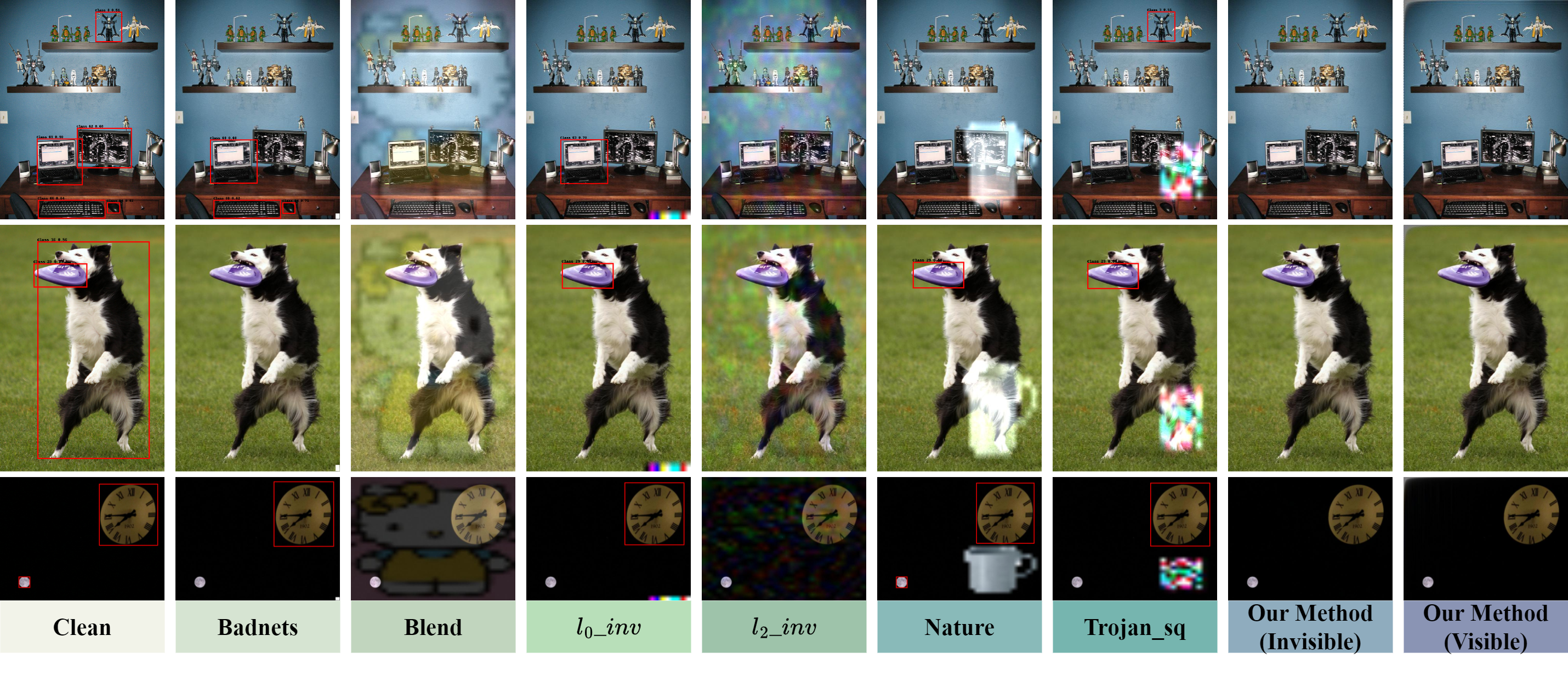}
   \caption{Comparison of visual results between our method and other backdoor attack methods.}
   \label{fig:results}
\end{figure*}

\begin{figure*}[t]
  \centering
   \includegraphics[width=1\linewidth]{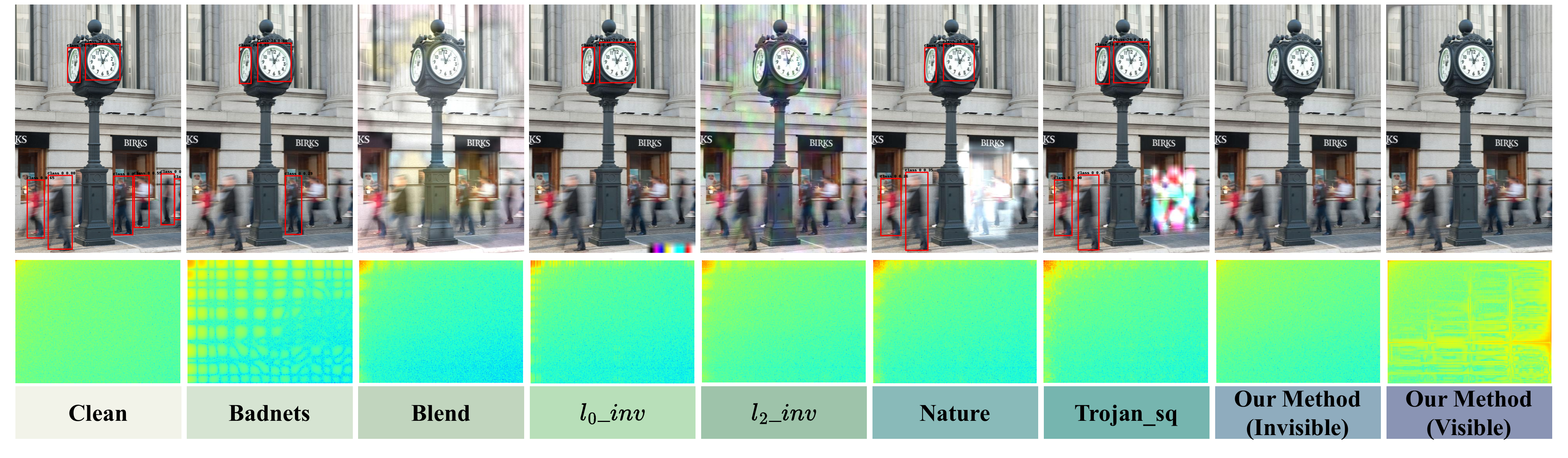}
   \caption{Comparisons of both the visual detection results (first row) and corresponding frequency domain (second row) between our method and other backdoor attack methods. The invisible trigger we propose avoids exhibiting high-frequency artifact.}
   \label{fig:hf}
\end{figure*}

\textit{1) Definition of Shapley Value ~\cite{chen2022rethinking,shapley1953value}:} We convert the image into an image in the frequency domain through DCT transformation. Then, the image is divided into $n$ rows and $n$ columns (here $n = 2$) with $n^2$ patches. Given any patch $\mathcal{B}_{i,j}$ in row $i$ and column $j$, We do not consider it and enumerate all possibilities of the remaining patches: other patches either appear or do not appear (are masked), then we will get $2^{n^2 - 1}$ possible situations.

Considering the patch $\mathcal{B}_{i,j}$, each possible situation will correspond to two images: and image $X_k$ with the patch $\mathcal{B}_{i,j}$ not masked and image $Y_k$ with the patch $\mathcal{B}_{i,j}$ masked. The difference in area loss of all images $X_k$s and $Y_k$s obtained by the above method on the detection model $D$ will be used as the contribution of the block, that is, the Shapley value $\mathcal{S}(\mathcal{B}_{i,j})$ 
\eqn{\label{eqn:shapley}}{
\mathcal{S}(\mathcal{B}_{i,j}) = \frac{1}{N}\sum_{k = 1}^{N} [F(X_k) - F(Y_k)],
}
where $N = 2^{n^2 - 1}$ and $F(X)$ is a function defined on the image $X$ as in Eqn. (\ref{eqn:detection-loss-func}).  

Note that the calculation of the Shapley value in Eqn. (\ref{eqn:shapley}) has exponential time complexity. In our experiments, we randomly sample $M$ samples from $2^{n^2 - 1}$ possibilities to obtain an estimate of the Shapley value
\eqn{}{
\widetilde{\mathcal{S}}(\mathcal{B}_{i,j}) = \frac{1}{M}\sum_{k = 1}^{M} [F(X_k) - F(Y_k)]. 
}

\textit{2) Analysis of Sharpley Value:} We compare the contribution of each image triplet (clean image, invisibly poisoned image, visibly poisoned image) in the frequency domain block, where the Shapley value \cite{shapley1953value} of the image in each frequency domain block is normalized to a probability distribution for fair comparison. Finally, we can get the average Shapley value probability distribution of the three types of images in different frequency domain blocks, as shown in Fig. \ref{fig:shapley-value}. 

Fig. \ref{fig:results} gives a qualitative comparison of the attack effects of these methods on three images. The results suggest that 1) The invisible and visibly poisoned images generated by our trigger generative networks have very similar attack effects. 2) The victim object detector performs abnormal behaviour on all invisible and visibly poisoned images generated by our trigger generative networks, and is better than other backdoor attack methods. 3) The victim object detector performs normal behaviour on clean images. 

Additionally, we transform all the images into the frequency domain for comparison, as shown in Fig. \ref{fig:hf}. The results indicate that the invisibly poisoned images generated by our proposed TGN1 do not exhibit high-frequency artifacts. Similar to the original images, they are smooth in the frequency domain without outliers. This also shows that the invisibly poisoned images achieve invisibility in both the frequency domain and the spatial domain, significantly enhancing the stealthiness of our method.

\section{Conclusion} \label{conclu}

We propose a general framework for training visible and invisible trigger generative networks against any object detector. Furthermore, our method can use invisible triggers in the training phase and visible triggers in the inference phase, making the attack process difficult to trace, thus achieving more stealthy backdoor attacks. The invisible trigger generative network incorporates a Gaussian smoothing layer and a high frequency artifacts classifier to optimize the stealthiness in both frequency and spatial domains, and the visible trigger generative network creates visible triggers equivalent to the invisible ones by employing alignment loss. Extensive experiments indicate that our method achieves superior attack performance in both visible and invisible trigger scenarios compared to other methods, and analyses are presented to investigate the inconsistency between invisible trigger and visible trigger. The proposed framework will inspire more investigations of the backdoor learning mechanism.

It is worth mentioning that the most important contribution of our attack algorithm is to improve the security of the object detection algorithm. For our attacks with visible triggers or invisible triggers, we can confirm whether the image contains harmful triggers by detecting abnormal pixels in the image in the spatial domain or in the frequency domain, respectively.

\bibliographystyle{IEEEtran}
\bibliography{main}

\end{document}